\documentclass[10pt, a4paper]{article}
\usepackage{lrec2022} 
\usepackage{multibib}
\usepackage{comment}
\newcites{languageresource}{Language Resources}
\usepackage{graphicx}
\usepackage{adjustbox}
\usepackage{amsmath}
\usepackage{float}
\usepackage{tabularx}
\usepackage{multicol}

\usepackage{soul}
\usepackage{titlesec}
\titleformat{\section}{\normalfont\large\bfseries\center}{\thesection.}{1em}{}
\titleformat{\subsection}{\normalfont\SmallTitleFont\bfseries\raggedright}{\thesubsection.}{1em}{}
\titleformat{\subsubsection}{\normalfont\normalsize\bfseries\raggedright}{\thesubsubsection.}{1em}{}
\renewcommand\thesection{\arabic{section}}
\renewcommand\thesubsection{\thesection.\arabic{subsection}}
\renewcommand\thesubsubsection{\thesubsection.\arabic{subsubsection}}

\usepackage{epstopdf}
\usepackage[utf8]{inputenc}

\usepackage{hyperref}
\usepackage{xstring}

\usepackage{xcolor}

\newcommand{\bs}[1]{{\color{orange}BS: #1}}

\title{Is something better than nothing?\\A story of Zero Shot Cross-lingual versus Unsupervised keyword detection \bs{kaj sta something/nothing tu?}}
\title{\bs{morda tudi: }Out of thin air: Zero-shot cross-lingual keyword detection}

\title{ Out of Thin Air: Is Zero-Shot Cross-Lingual Keyword
Detection Better Than Unsupervised?}

\name{Boshko Koloski, Senja Pollak, Bla\v{z} \v{S}krlj, Matej Martinc} 

\address{Jo\v{z}ef Stefan Institute, Jo\v{z}ef Stefan International Postgraduate School \\
         Jamova cesta 39, Ljubljana, Slovenia \\
         \{boshko.koloski,senja.pollak,blaz.skrlj,matej.martinc\}@ijs.si\\
        }

\abstract{
Keyword extraction is the task of retrieving words that are essential to the content of a given document. Researchers proposed various approaches to tackle this problem. At the top-most level, approaches are divided into ones that require training - supervised and ones that do not - unsupervised.  In this study, we are interested in settings, where for a language under investigation, no training data is available. More specifically, we explore whether pretrained multilingual language models can be employed for zero-shot cross-lingual keyword extraction on low-resource languages with limited or no available labeled training data and whether they outperform state-of-the-art unsupervised keyword extractors. The comparison is conducted on six news article datasets covering two high-resource languages, English and Russian, and four low-resource languages, Croatian, Estonian, Latvian, and Slovenian. We find that the pretrained models fine-tuned on a multilingual corpus covering languages that do not appear in the test set (i.e. in a zero-shot setting), consistently outscore unsupervised models in all six languages.
\newline \Keywords{keyword detection, cross-lingual learning, zero-shot learning} 
 }

\begin{document}

\maketitleabstract

\section{Introduction}
Detecting keywords represents a crucial task in several text intensive applications. News industry relies on keywords for organization, linking and summarization of articles according to the content and topics they cover. With the current trend of fast-paced type of writing and an ever-growing amount of generated news, it becomes an infeasible task for the journalists to manually extract keywords and the development of tools for automatic extraction has become essential for speeding up the media production. 

Keyword extraction can be tackled in a supervised or an unsupervised way. The current supervised state-of-the-art approaches are based on transformer-based \cite{vaswani2017attention} deep neural networks and employ large-scale language model pretraining. Despite being very successful in solving the task, they do require substantial amounts of labeled data which is expensive to obtain or non-existent for some low-resource languages and domains. To cope with this, researchers in most cases employ unsupervised keyword extraction in this low-resource scenarios. Unsupervised approaches require no prior training and can be applied to most languages, making them a perfect fit for domains and languages that have low to no amount of labeled data. On the other hand, they offer non-competitive performance when compared to supervised approaches \cite{martinc2020tnt}, since they can not be adapted to the specific language, domain and keyword assignment regime through training.

In this work, we explore another option for keyword extraction in low-resource settings, which has not been extensively explored in the past, a zero-shot cross-lingual keyword detection. More specifically, we investigate how multilingual pretrained language models, which have been fine-tuned to detect keywords on a set of languages, perform, when applied to a new language not included in the train set, and compare these results to the results achieved by several state-of-the-art unsupervised keyword extractors. In addition, we also investigate whether in a setting, where training data is available, supervised monolingual models can benefit from additional data from another language. The main contributions are the following:

\begin{itemize}
    \item 
    We conduct and extensive zero-shot cross-lingual study of keyword extraction on six languages, four of them less-resourced European languages, and demonstrate that a multilingual BERT model fine-tuned on the training data not matching target language, performs better than state-of-the-art unsupervised keyword extraction algorithms. 
    \item We evaluate the performance of supervised zero-shot cross-lingual models in comparison to the supervised monolingual models in order to better determine the decrease in the performance when no language specific data is available.
    \item We investigate if the performance of monolingual models can be improved by including additional multilingual data and whether there is a trade-off between the amount of data available and the language specificity of this data.
    \item We produce new supervised keyword extraction models for a new \textit{Slovenian} dataset for keyword extraction, contributing to the development of new language resources for a less-resourced European language. 
\end{itemize}

The rest of this paper is organized in the following way: Section \ref{sec:related_work} presents the related work in the field of keyword extraction, focusing also on the cross-lingual zero-shot learning. Section \ref{sec:data} describes the data used in our experiments and Section \ref{sec:experiments} explains our experimental settings. While Section \ref{sec:discusion} presents and discuses the results of our experiments, Section \ref{sec:conclusion} concludes the paper and proposes further work on this topic.

\section{Related Work}
\label{sec:related_work}
We can divide approaches for keyword extraction into supervised and unsupervised. As stated above, state-of-the-art supervised learning approaches have become very successful at tackling the keyword extraction task but are data-intensive and time consuming. Unsupervised keyword detectors can tackle these two problems and usually require a lot less computational resources and no training data, yet this comes at the cost of the reduced overall performance.

We can divide unsupervised approaches into four main categories, namely statistical, graph-based, embeddings-based, and 

language model-based methods. Statistical and graph based methods are the most popular and the main difference between them is that statistical methods, such as KPMiner \cite{el2009kp}, RAKE \cite{rose2010automatic}, and YAKE \cite{yake}, leverage various text statistics to capture keywords, while Graph-based methods, such as TextRank \cite{mihalcea2004textrank}, Single Rank, KeyCluster \cite{liu2009clustering}, and RaKUn \cite{DBLP:journals/corr/abs-1907-06458} build graphs and rank words according to their keyword potential based on their position in the graph. Among the most recent statistical approaches is YAKE \cite{yake}, which we also test in this study. It is based on features such as casing, position, frequency, relatedness to context and dispersion of a specific term, which are heuristically combined to assign a single score to every keyword. KPMiner \cite{el2009kp} is an older, simpler method that focuses on the frequency and the position of appearance of a potential keyphrase. In order to enrich the quality of the retrieved phrases, the model proposes several filtering steps, e.g. removing rare candidate phrases that do not appear at least \textit{n}-times and that do not appear within some cutoff distance from the beginning of the document. 

TextRank \cite{mihalcea2004textrank}, which we evaluate in this study, is one of the first graph-based methods for keyword detection. It leverages Google's PageRank algorithm to rank vertices in the lexical graph according to their importance inside a graph. Other method that employs PageRank is PositionRank \cite{florescu2017positionrank}. The so-called MultiPartiteRank algorithm \cite{Multipartite} encodes the potential candidate keywords of a given document into a multipartite graph structure, which also considers topic information. In this graph two nodes, representing keyphrase candidates, are connected only if they belong to different topics and the edges are weighted according to the distance between
the two candidates in the document. In order to rank the verticies, the method leverages PageRank, similarly to \newcite{mihalcea2004textrank}. One of the most recent graph-based keyword extractors is RaKUn \cite{DBLP:journals/corr/abs-1907-06458}. The main novelty in this algorithm is the expansion of the initial lexical graph with the introduction of meta-vertices, i.e., aggregates of existing vertices. It employs \textit{load centrality} measure for ranking vertices and relies on several graph redundancy filters.  

Embedding-based keyword extraction methods are less popular but are nevertheless recently gaining traction. The first methods of this type were proposed by \newcite{wang2014corpus}, who proposed Key2Vec \cite{mahata2018key2vec}, and \newcite{bennani-smires-etal-2018-simple}, who proposed EmbedRank. Both of these methods employ semantic information from distributed word and sentence representations. The most recent state-of-the-art method of this type is KeyBERT proposed by \newcite{grootendorst2020keybert}, which leverages pretrained BERT based embeddings for keyword extraction. In this approach, embedding representations of candidate keyphrases are ranked according to the cosine similarity to the embedding of the entire document. 

Language model-based keyword methods, such as the ones proposed by \newcite{tomokiyo2003language} use language model derived statistics to extract keywords from text. These type of keyword extraction models are quite rare and are not included in our study.

One of the first supervised approaches to keyword extraction was KEA proposed by \newcite{witten2005kea}. It considers keyword identification as a classification task and employs Naive Bayes classifier to determine for each word or phrase in the text if it is a keyword or not. It uses only TF-IDF and the term's position in the text as classification features. A more recent non-neural supervised approach employs a sequence labelling approach to keyword extraction and was proposed by \newcite{gollapalli2017incorporating}. The approach relies on Conditional Random Field (CRF) tagger. First neural sequence labeling approach was proposed by \newcite{luan2017scientific}, who proposed a neural network comprising of a bidirectional Long Short-Term Memory (BiLSTM) layer and a CRF tagging layer. 

Keyword detection can also be considered as a sequence-to-sequence generation task. This idea was first proposed by \newcite{meng2017deep}, who employed a recurrent generative mode with an attention mechanism and a copying mechanism \cite{gu2016incorporating} based on positional information for keyword prediction. What distinguishes this model from others is that besides being able to detect keywords in the input text sequence, it can also potentially find keywords that do not appear in the text.  

The most recent approaches tackle keyword detection with transformer architectures \cite{vaswani2017attention} and formulate keyword extraction task as a sequence labelling task. In the study by \newcite{sahrawat2019keyphrase}, contextual embeddings generated using BERT \cite{devlin2019bert}, RoBERTa \cite{liu2019roberta} and GPT-2 \cite{radford2019gpt2} were fed into a bidirectional Long short-term memory network (BiLSTM) with an optional Conditional random fields layer (BiLSTM-CRF). They conclude that contextual embeddings generated by transformer architectures outperform static. Another study employing transformer architecture and sequence labelling approach was conducted by \newcite{martinc2020tnt}. Their approach, named TNT-KID did not rely on massive pretraining but rather on pretraining the transformer based language model on much smaller domain specific corpora. They report good results employing this tactic and claim that this makes their model more transferable to low-resource languages with limited training resources.

Most keyword detection studies still focus on English. Nevertheless, recently several multilingual and cross-lingual studies, which also include low-resource languages, were conducted. On of them is the study by \newcite{koloski-etal-2021-extending} where the performance of two supervised transformer-based models, multilingual BERT with a BiLSTM-CRF classification head and TNT-KID were compared in a multilingual settings, on Estonian, Latvian, Croatian and Russian news corpora. The authors also explored if combining the outputs of the supervised models with the outputs of unsupervised models can improve the recall of the system. 

Cross-lingual zero-shot transfer represents an arising hot-topic in the research community. The main idea behind this family of approaches is that models can benefit from transfer from one language to another and therefore be able to conduct tasks in new, `unseen languages`, on which they were not trained in a supervised way. These approaches are especially useful for low-resource languages without manually labeled resources. We are aware of two unsupervised cross-lingual approaches to keyword extraction. One of them is BiKEA \cite{huang2014cross}, where the authors construct word graphs for documents in parallel corpora and rely on cross-lingual word statistics for keyword extraction. Another one is the study by \newcite{cross_ling_latent}, where the focus is on building single latent space over two languages, and later extracting keywords, to be used as topic categories for the articles, from this common latent space. 

Researchers conducted various studied on the effect of applying zero-shot cross-language modelling to multiple domains of NLP, with most of the experiments showing promising results. For example, a zero-shot approach, in which a model was trained on one language and applied on the other, for the task of automatic reading comprehension was carried out by \newcite{crossling_comperhension}. Phoneme recognition is another task that cross-lingual zero-shot learning seems to improve. In the work by \newcite{crossling_phoeme} they show that cross-lingual phoneme recognition offers performance comparable to the state-of-the-art unsupervised models for the task at hand. 

Recently, masked language models based on transformers such as BERT \cite{devlin2019bert} have taken the field by the storm, achieving state-of-the-art results on many tasks. In a study by \newcite{crossling_bert_study} they explored how well does the multilingual variant of BERT performs when used in a zero-shot setting. The study included 39 languages and covered 5 different tasks, including document classification, natural language inference, named entity recognition, part-of-speech tagging, and dependency parsing. The results were very promising, with the model outscoring several unsupervised and non-transformer based cross-lingual approaches. A zero-shot approach relying on multilingual BERT was also adopted to tackle the tasks of news-sentiment classification \cite{app10175993}, offensive speech detection \cite{pelicon2021investigating} and abusive language detection \cite{glavas-etal-2020-xhate}. These studies concluded that pretrained models can be used in a cross-lingual fashion, serving as a strong baseline in the low-resource scenario. To the best of our knowledge, zero-shot transfer has not yet been investigated for the task of keyword extraction.

\section{Data}
\label{sec:data}
\begin{table*}[!h]
    \centering

   \resizebox{\textwidth}{!}{ 

    \begin{tabular}{|c|c|c|c|c|c|c|c|c|c|} 
          \cline{2-10}
         \multicolumn{1}{c|}{}& \multicolumn{3}{c|}{ \textbf{Train}} & \multicolumn{3}{c|}{\textbf{Valid}} & \multicolumn{3}{c|}{\textbf{Test}}  \\ \hline
 Language &     size &   kw\_per\_doc &  kw\_present &    size &   kw\_per\_doc &  kw\_present &    size &   kw\_per\_doc &  kw\_present \\ \hline
       Latvian &  10506  &    3.2204 &      0.8691 &  2627  &       3.2687 &      0.8658 & 11641  &      3.1964 &      0.8624 \\ \hline
 Estonian &   8600  &    3.8244 &      0.7809 &  2150  &     3.7386 &      0.7785 &  7747  &       4 944 &      0.8073 \\ \hline
Slovenian &   4796  &    4.0052 &      0.5991 &  1199  &     4.1643 &      0.6054 &  1519  &      3.8861 &      0.5995 \\ \hline
 Croatian &  25778  &    3.5375 &      0.7047 &  6445  &     3.5469 &      0.6988 &  3582  &       3.5274 &      0.7009 \\ \hline
  English & 207938  &  5.324 &      0.4599 & 51985  &     5.0350 &      0.4583 & 20000  &       5 349 &      0.6205 \\ \hline
  Russian &  11064  &    5.6377 &      0.7779 &  2767.0 &    5.7311 &      0.7797 & 11475.0 &   5.4261 &      0.7918  \\ \hline
    \end{tabular} 
    }
    \caption{Number of documents (size), keywords per document (kw\_per\_doc) and percentage of keywords present in document's text (kw\_present) per split in our experiments. Percentage of present keywords represents the percentage of keywords that appear in the text of the document.}
    \label{tab:datasets}

\end{table*}

For model evaluation we use six different datasets from the news domain. 
We include Russian, Croatian, Latvian, and Estonian news article datasets with manually labeled keywords from the
\newcite{clarin_kw} dataset repository, using the same splits as in 
\newcite{koloski-etal-2021-extending}. Additionally, we include a benchmark English dataset, the KPTimes dataset \cite{gallina2019kptimes}, and a Slovenian SentiNews  \cite{slovenian_dataset}, which was originally used for news sentiment analysis, but nevertheless does contain manually labeled keywords and was therefore identified as suitable for keyword extraction. Before feeding the datasets to the models, they are lowercased. Each dataset is split into three different splits: \textit{train}, \textit{validation} and \textit{test}. For English, we use the data splits introduced in \cite{gallina2019kptimes}, for other languages besides Slovenian we use the same data splits as in \cite{koloski-etal-2021-extending}, while for Slovene we first removed the articles without keywords and randomly split the dataset into training, validation and test splits. We use the splits in the following manner:
\begin{itemize}
    \item \textit{train split} - used for fine-tuning of the cross-lingual supervised model. The procedure is explained in detail in Section \ref{subsec:supervised}.
    \item \textit{valid split} - used for early stopping in order to prevent over-fitting during the fine-tuning phase of the supervised models.
    \item \textit{test split} - used for evaluation of the supervised and unsupervised methods. This split is not used during training of any of the methods.
\end{itemize}
 The dataset statistics are available in Table \ref{tab:datasets}. For each split we report on the \textit{size} (number of articles), the average amount of keywords per document (\textit{kw\_per\_doc}) and finally the percentage of keywords that actually appear in the text of the news articles (\textit{kw\_present}). \textit{Latvian} dataset has on average least keywords per document ($3.22$) while the English and Russian datasets contain most keywords per article, $5.32$ and $5.64$, respectively. 
 
 Note that some of the keywords accompanying an article in the data do not appear in the text of the document. For evaluation purposes we only use the \textbf{keywords present} in the documents. \textit{English} has the lowest amount of present keywords ($46\%$), while \textit{Latvian} has the highest percentage of present keywords ($87\%$). We consider keyword or keyphrase as present if a stemmed (English and Lativan) or lemmatized version (for other languages) appears in the stemmed or lemmatized version of the document. We use the NLTK's \cite{bird2009natural} implementation of the \textit{PorterStemmer} for English and \textit{LatvianStemmer}\footnote{\url{https://github.com/rihardsk/LatvianStemmer}} for Latvian. For \textit{Croatian, Slovenian, Estonian} and \textit{Russian} we use the \textit{Lemmagen3} \cite{jurvsic2010lemmagen} lemmatizer.

\section{Experimental Setup}
\label{sec:experiments}
In our experiments, we employ several unsupervised models to which we compare several supervised cross-lingual, multilingual and monolingual approaches.

\subsection{Unsupervised Approaches}
\label{subsec:unusper}
We evaluate three types of unsupervised keyword extraction methods, statistical, graph-based, and embedding-based, described in Section \ref{sec:related_work}.

\subsubsection{Statistical Methods} 

\begin{itemize}
\item \textbf{YAKE} \cite{yake}: We consider n-grams with $n \in \{1,2,3\}$ as potential keywords. 
\item \textbf{KPMiner} \cite{el2009kp}: We apply least allowable seen frequency of $3$, while we set the \textit{cutoff} to $400$.
\end{itemize}

\subsection{Embedding-based Methods}
\begin{itemize}

\item \textbf{KeyBERT} \cite{grootendorst2020keybert}: For document embedding generation we employ sentence-transformers \cite{reimers-2019-sentence-bert}, more specifically the \textit{distiluse-base-multilingual-cased-v2} model available in the Huggingface library\footnote{\url{https:/huggingface.co/sentence-transformers/distiluse-base-multilingual-cased-v2}}. Initially, we tested two different KeyBERT configurations: one with n-grams of size $1$ and another with n-grams ranging from 1 to 3, with \textit{MMR}=$false$ and with \textit{MaxSum}=$false$. The unigram model outscored the model that considered n-grams of sizes 1 to 3 as keyword candidates for all languages, therefore in the final report we show only the results for the unigram model.

\end{itemize}

\subsubsection{Graph-based Methods} 

\begin{itemize}
    \item \textbf{TextRank} \cite{mihalcea2004textrank}: For languages supported by the PKE library \cite{pke} (Russian and English), we employ stemming for normalization, and part-of-speech tagging during candidate weighting. $33\%$ of the highest ranked words are considered as potential candidates. 
    \item \textbf{MultipartiteRank} \cite{Multipartite}: We employ part-of-speech tagging during candidate weighting for supported languages, and we set the minimum similarity threshold for clustering at $74\%$. 
    \item \textbf{RaKUn} \cite{DBLP:journals/corr/abs-1907-06458}: We use edit distance for calculating distance between nodes, use language specific stopwords from the \textit{stopwords-iso} library\footnote{\url{https://github.com/stopwords-iso/stopwords-iso}}, a \textit{bigram-count\_threshold} of $2$ and a \textit{distance\_threshold} of $2$.
\end{itemize}

We use the PKE \cite{pke} implementations of 
\textit{YAKE}, \textit{KPMiner}, \textit{TextRank} and \textit{MultiPartiteRank}.  We use the official implementation for the RaKUn model \cite{DBLP:journals/corr/abs-1907-06458} and for the KeyBERT model \cite{grootendorst2020keybert}. For unsupervised models, the number of returned keywords need to be set in advance. Since we employ F1@10 as the main evaluation measure (see Section \ref{sec:evaluation-setting}), we set the number of returned keywords to 10 for all models.

\subsection{Supervised Approaches}
\label{subsec:supervised}
We utilize the transformer-based BERT model \cite{devlin2019bert} with a token-classification head consisting of a simple linear layer for all our supervised approaches. We treat the keyword extraction task as a sequence classification task. We follow the approach proposed in \newcite{martinc2020tnt} and predict binary labels (1 for `keywords' and 0 for `not keywords') for all words in the sequence. The sequence of two or more sequential keyword labels predicted by the model is always interpreted as a multi-word keyword. We do not follow the related work \cite{koloski-etal-2021-extending} on adding a BilSTM-CRF classification head on top of BERT for sequence classification. Sincethe  classification head needs to be randomly initialized (i.e. it was not pretrained during the BERT pretraining) and since, among others, we apply the model in a cross-lingual setting, we prefer to keep the token classification head simple, since the layers inside the head do not obtain any multilingual information during fine-tuning. The hypothesis is that using a simple one-layer classification head will result in a better generalization of the model in a cross-lingual setting.

More specifically, we employ the \textit{bert-uncased-multilingual} model from the HuggingFace library \cite{huggingface} and fine-tune it using an adaptive learning rate (starting with the learning rate of $3\cdot10^{-5}$), for up to $10$ \textit{epochs} with a batch-size of $8$.

\subsubsection{Cross-lingual Setup}

Let $C_{k}$ be the collection of all of the possible tuples of size $k$ that can be constructed from the $6$ languages. For example, $C_{2}$ denotes the collection of all possible two language combinations in a set of $6$ languages, e.g. 
\begin{equation*}
    C_{2} = \{(English, Russian), (English, Latvian) , \dots \}
\end{equation*} 
We denote the \textit{i-th} tuple of size $k$ with $C_{k}^{i}$, e.g. for the previous example, $C_{2}^{1}$ would yield $(English, Russian)$. The cardinality of the collection $C_{k}$, $|C_{k}|$ is calculated as:
\begin{equation*}
    |C_{k}| = \binom{6}{k}
\end{equation*}

We create the \textit{i-th} training dataset D from the collection of tuples $C_{k}$ of size \textit{k}, as a concatenation of datasets in the tuple, or more formally $D_{i,k}$:
\begin{equation*}
       D_{i,k} =  \bigcup_{language \in C_{k}^{i}} \textit{train-split(language)} 
\end{equation*}
where \textit{train-split} represents the respective data-split of the given \textit{language} as described in Section \ref{sec:data}.

Dependent on the number of languages $k$ included in the training set, and depending on what languages are the trained models employed, we define the following specific settings, for which we report results in Section \ref{sec:discusion}:
\begin{itemize} 
    \item \textbf{MON} - monolingual ($k=1$ ; $D_{i,1}$ for $i \in {1,|C_{1}|}$) - where we fine-tune the model on a single language (for example \textit{English}). In this setting we train a total of 6 \textbf{monolingual} models\footnote{Note that even in this `monolingual setting' we employ BERT pretrained on a multilingual corpus, since we are more interested in the comparison of fine-tuning regimes in this paper than in the comparison of different pretrained models.} and \textbf{we train and test each model on the same language}. We use this setting as a baseline to which we compare unsupervised, cross-lingual and multilingual settings, i.e. for cross-lingual (LOO) and unsupervised settings, MON indicates how much we would gain, if language specific training data was available.
    \item \textbf{LOO} - \textit{Leave One Out} ($k=5$ ; $D_{i,5}$ for $i \in {1,|C_{5}|}$) - where we fine-tune the model on a concatenation of five languages (for example \textit{Slovenian, Estonian, Latvian, Russian, Croatian}) and test it on the sixth language not appearing in the train set (e.g. \textit{English}). In this manner we obtain 6 different models. This is the so-called \textbf{zero-shot cross-lingual} setting, since we do not include the test language at the training time. The main idea behind this setting is to test how well does a model do if no language specific training data is available. This setting represents the core of our experiments.
    \item \textbf{MUL} - multilingual ($k=6$ ; $D_{i,6}$ for $i \in {1,|C_{6}|}$) - where we fine-tune just one model on all languages from the language set and apply it on all the test datasets. With this experiment we want to check if adding more domain-specific data from other languages improves the performance in comparison to the monolingual setting described above. 
\end{itemize}

\subsection{Evaluation Setting}
\label{sec:evaluation-setting}

In order to evaluate the models, we calculate F1, recall and precision at 10 retrieved words. We omit the documents that do not have present keywords or do not contain keywords. We do this since we only use approaches that extract words (or multi-word expressions) from the given document and cannot handle keywords not appearing in the text. All approaches are evaluated on the same monolingual test splits, which are not used for training of supervised models. Lowercasing and stemming (for English and Latvian) or lemmatization (for other languages) are performed on both the gold standard and the extracted keywords (keyphrases) during the evaluation. 
\section{Discussion of Results}
\label{sec:discusion}

Table \ref{tab:evaluation_F1} presents the results in terms of F1@10, Table \ref{tab:evaluation_precision} presents the results in terms of precision@10 and Table \ref{tab:evaluation_recall} presents the results in terms of recall@10.

All unsupervised approaches are outperformed by the cross-lingual approaches (see row LOO) across all of the datasets and according to all criteria. For all languages besides Slovenian, the cross-lingual model improves on the best performing unsupervised model by more than 10 percentage points in terms of F1@10, the improvement being the smallest for Slovenian (about 8 percentage points) and the biggest for Latvian and Estonian (about 15 percentage points). The best performing unsupervised model in terms of F1@10 is KeyBert, which outperforms other unsupervised models on all languages.

The difference in F@10 between the cross-lingual and monolingual models (see row MON) is substantial. If no training data for the target language is used, the performance is more than halved on three languages, Latvian, Estonian and Russian. For the other three languages, the drop is smaller yet still substantial. Similar drops can be observed according to two other measurements precision@10 and recall@10.

The monolingual and multilingual models offer comparable performance according to all measures and across all languages. This indicates that including other languages into the train set, besides the target language, does generally not improve performance of the models, especially if the training dataset is sufficiently large. This finding supports the so-called curse of multilinguality \cite{conneau2019unsupervised}, i.e. a trade-off between the number of languages the model supports and the overall decrease in performance on monolingual and cross-lingual benchmarks. It is however very likely that the transfer between languages would be more successful if the language set would contain more similar languages.

\begin{table}[!ht]
\resizebox{\linewidth}{!}{\begin{tabular}{|c|c|c|c|c|c|c|c|}
        \hline
    \multicolumn{2}{|c|}{Language} & English & Slovenian  & Croatian   & Latvian   & Estonian   & Russian   \\ \hline
    Model  & T & \multicolumn{6}{c|}{F1@10} \\ \hline
    \hline
\multicolumn{8}{|c|}{Without training data in the target language} \\ \hline

 KPMiner&U&0.1584&0.0941&0.1043&0.131&0.0641&0.0578 \\ \hline 
 YAKE&U&0.1449&0.0794&0.1248&0.095&0.0653&0.0966 \\ \hline
 KeyBert&U&0.1702&0.1153&0.1668&0.1330&0.0923&0.1352\\ \hline  
 TextRank&G&0.0440&0.0042&0.0041&0.0196&0.0239&0.0392 \\ \hline 
 RaKUn&G&0.1176&0.0875&0.0902&0.0862&0.0605&0.0731 \\ \hline  
 MPRU&G&0.1549&0.0455&0.0683&0.0821&0.0398&0.1171 \\ \hline    
LOO  &  C &\textbf{0.2856}&\textbf{0.2000}&\textbf{0.2883}&\textbf{0.2844}&\textbf{0.2368}&\textbf{0.2395} \\ \hline \hline
\multicolumn{8}{|c|}{With training data} \\ \hline
MON & S & 0.4658 & 0.3259 & 0.4644 &\textbf{ 0.6533 } & \textbf{ 0.4920 } & \textbf{ 0.5979 } \\ \hline
MUL & S &\textbf{0.4702}&\textbf{0.3371}&\textbf{0.4674}&0.6532&0.4900&0.5943 \\ \hline    
    \end{tabular}}

    \caption{Performance of the models according to the F1@10. The \textit{T} column denotes the type of model - \textit{U} denotes unsupervised statistical model, \textit{G} denotes unsupervised graph based model, \textit{S} denotes the supervised BERT model and finally \textit{C} denotes the cross-lingual 
    \textit{LOO} model. MPRU entry in the Model column denotes the MultiPartiteRank model.}
    \label{tab:evaluation_F1}
\end{table}

\begin{table}[ht!]
\resizebox{\linewidth}{!}{ \begin{tabular}{|c|c|c|c|c|c|c|c|}
        \hline
    \multicolumn{2}{|c|}{Language} & English & Slovenian  & Croatian   & Latvian   & Estonian   & Russian   \\ \hline
    Model  & T & \multicolumn{6}{c|}{precision@10}\\ \hline
    \hline
\multicolumn{8}{|c|}{Without training data in the target language} \\ \hline
 KPMiner&U&0.1493&0.1280&0.0974&0.1243&0.0822&0.0578 \\ \hline 
 YAKE&U&0.1068&0.0591&0.0818&0.0602&0.0432&0.0966 \\ \hline
 KeyBert&U&0.1640&0.1213&0.1428&0.0995&0.0747&0.1352\\ \hline  
 TextRank&G&0.0322&0.0036&0.0028&0.0120&0.0157&0.0392 \\ \hline 
 RaKUn&G&0.0871&0.0672&0.0605&0.0550&0.0417&0.0731 \\ \hline  
 MPRU&G&0.1151&0.0339&0.0462&0.0524&0.0273&0.1171 \\ \hline    
LOO  &  C & \textbf{0.3337}&\textbf{0.2728}&\textbf{0.2955}&\textbf{0.3158}&\textbf{0.3247}&\textbf{0.2395} \\ \hline \hline
\multicolumn{8}{|c|}{With training data} \\ \hline
MON & S & 0.5278 & 0.2954 & 0.4514 & \textbf{0.7056} & 0.5053 & \textbf{ 0.5979 } \\ \hline
MUL & S & \textbf{0.5318}&\textbf{0.3429}&\textbf{0.4799}&0.7021&\textbf{0.5212}&0.5943 \\ \hline 
    \end{tabular}}

    \caption{Performance of the models according to the precision@10. The \textit{T} column denotes the type of model - \textit{U} denotes unsupervised statistical model, \textit{G} denotes unsupervised graph based model, \textit{S} denotes the supervised BERT model and finally \textit{C} denotes the cross-lingual 
    \textit{LOO} model. MPRU entry in the Model column denotes the MultiPartiteRank model. }
    \label{tab:evaluation_precision}
\end{table}

\begin{table}[ht!]
 \resizebox{\linewidth}{!}{\begin{tabular}{|c|c|c|c|c|c|c|c|}
        \hline
      \multicolumn{2}{|c|}{Language} & English & Slovenian  & Croatian   & Latvian   & Estonian   & Russian   \\ \hline
    Model  & T & \multicolumn{6}{c|}{recall@10} \\ \hline
    \hline
\multicolumn{8}{|c|}{Without training data in the target language} \\ \hline

 KPMiner&U&0.1688&0.0744&0.1123&0.1384&0.0525&0.0578 \\ \hline 
 YAKE&U&0.2251&0.1213&0.2625&0.2254&0.1336&0.0966 \\ \hline
 KeyBert&U&0.1768&0.1200&0.2001&0.2007&0.1206&0.1352\\ \hline  
 TextRank&G&0.0694&0.0051&0.0076&0.0536&0.0502&0.0392 \\ \hline 
 RaKUn&G&0.1813&0.1252&0.1772&0.1995&0.1099&0.0731 \\ \hline  
 MPRU&G&0.2367&0.0696&0.1310&0.1899&0.0734&0.1171 \\ \hline    
LOO  &  C &\textbf{ 0.2496}&\textbf{ 0.1579}&\textbf{ 0.2815}&\textbf{ 0.2586}&\textbf{ 0.1863}&\textbf{0.2395} \\ \hline \hline
\multicolumn{8}{|c|}{With training data} \\ \hline

MON & S & 0.4169 &\textbf{ 0.3634 } & \textbf{ 0.4781 } & 0.6082 &  \textbf{ 0.4794 } & \textbf{ 0.5979 } \\ \hline
MUL & S &\textbf{0.4215}&0.3314&0.4556&\textbf{0.6107}&0.4624&0.5943 \\ \hline         
    \end{tabular}}

    \caption{Performance of the models according to the recall@10. The \textit{T} column denotes the type of model - \textit{U} denotes unsupervised statistical model, \textit{G} denotes unsupervised graph based model, \textit{S} denotes the supervised BERT model and finally \textit{C} denotes the cross-lingual 
    \textit{LOO} model. MPRU entry in the Model column denotes the MultiPartiteRank model. }
    \label{tab:evaluation_recall}
\end{table}

\subsection{Adding More Languages in a Cross-lingual Setting}

\begin{figure}[H]
    \centering
     \includegraphics[width=\linewidth]{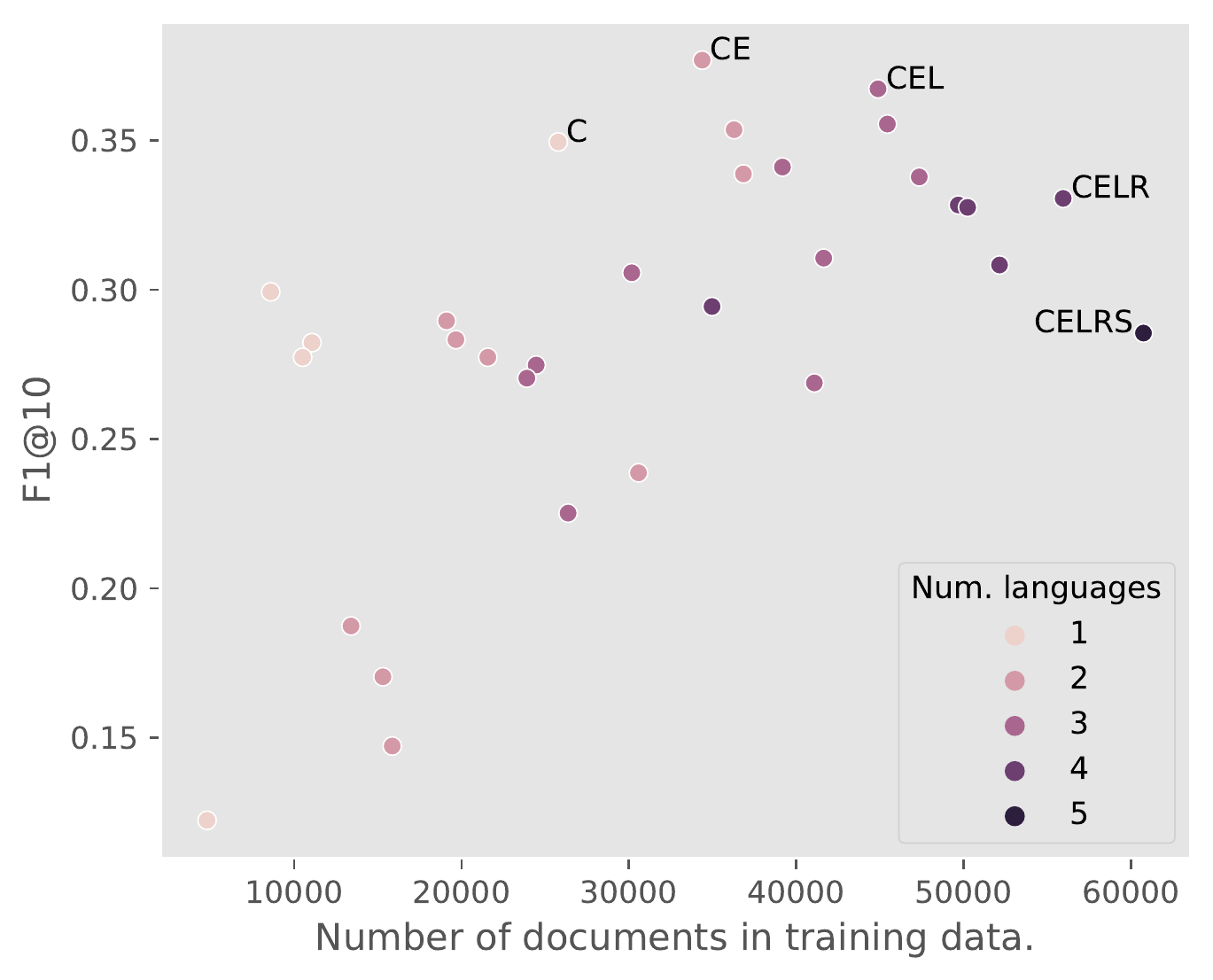}
    \caption{Correlation between the number of languages and the performance of the model according to the F1@10, when the model is tested on an unseen language (English). The best-performing combinations per language are labeled with a sequence of letters representing languages: Croatian - \textit{C}, Slovenian - \textit{S}, Estonian - \textit{E}, Latvian - \textit{L} and Russian - \textit{R}.}
    \label{fig:docsize_F1}
\end{figure}

\begin{figure}[H]
    \centering
    \includegraphics[width=\linewidth]{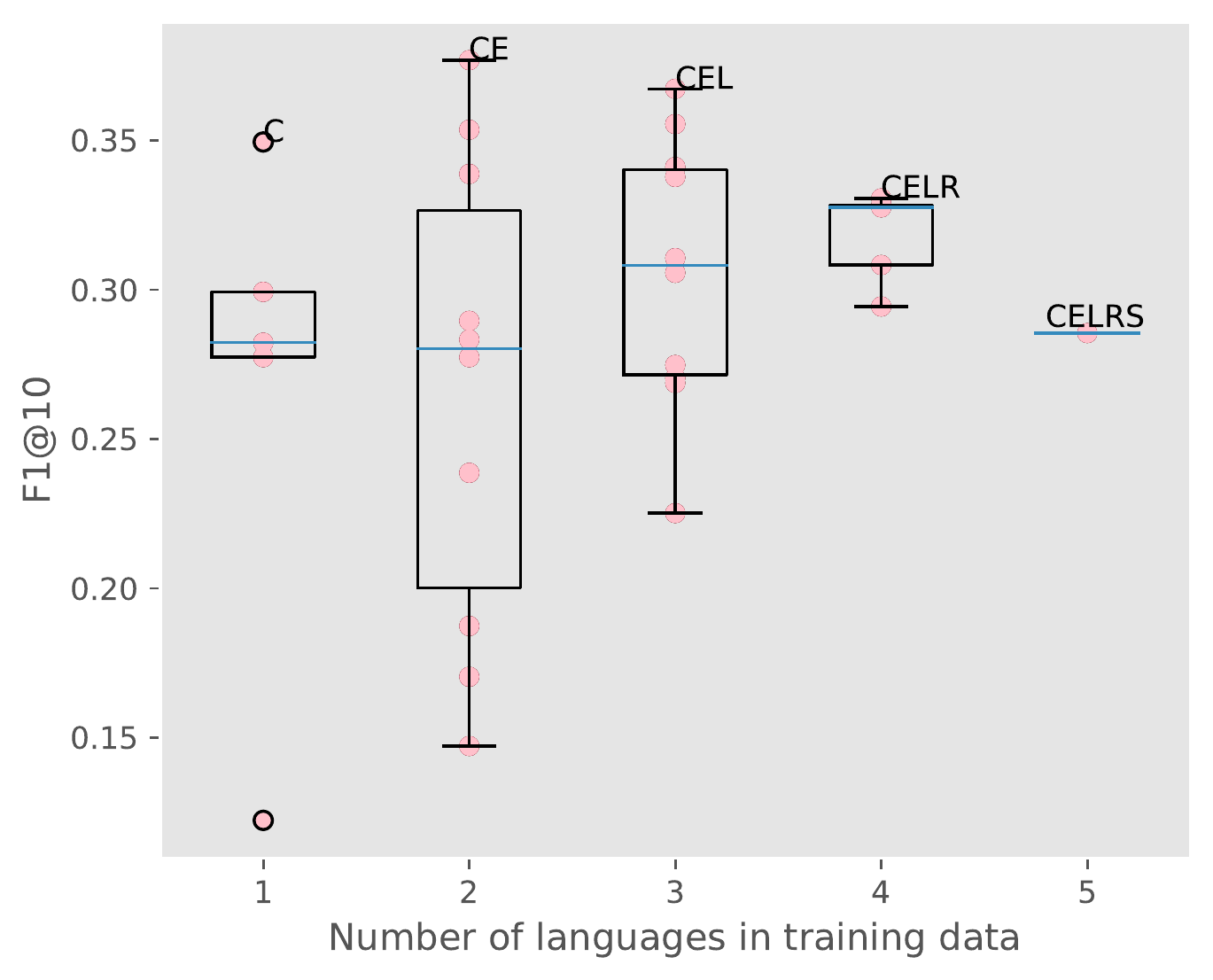}
    \caption{Correlation between the number of languages and the performance of the model according to the F1@10, when the models is tested on an unseen language (English). The models are split into groups according to the number of languages they were trained on and each group is represented by a boxplot. The best-performing combinations per language are labeled with a sequence of letters representing languages: Croatian - \textit{C}, Slovenian - \textit{S}, Estonian - \textit{E}, Latvian - \textit{L} and Russian - \textit{R}..}
    \label{fig:box_combined}
\end{figure}

Above we have showed that adding other languages into the train set already containing the data that matches the target language does generally not improve the performance. On the other hand, here we explore if it is worth adding more languages in a cross-lingual setting. We consider \textit{English} as a testing language, and train on different combinations of languages that do not include English. Figure \ref{fig:docsize_F1} presents the correlation between the number of languages and the performance of the model according to the F1@10. The Figure does indicate some positive correlation between the number of languages in the train set and the F1@10 improvement. The best was the model trained on Croatian (labeled as C) achieving F1@10 of $35\%$. Overall, the best performing model on English was trained on the concatenation of the Croatian and Estonian corpus (labeled as CE). Adding additional languages to the train set did not improve the performance further.  It does however improve the stability of the models, i.e. the models trained on more languages tend to have higher performance minimum but also lower performance maximum, as can be clearly seen in Figure \ref{fig:box_combined}.

\subsection{Zero-shot Performance of the Monolingual Models}

\begin{figure}[h]
    \centering
     \includegraphics[width=\linewidth]{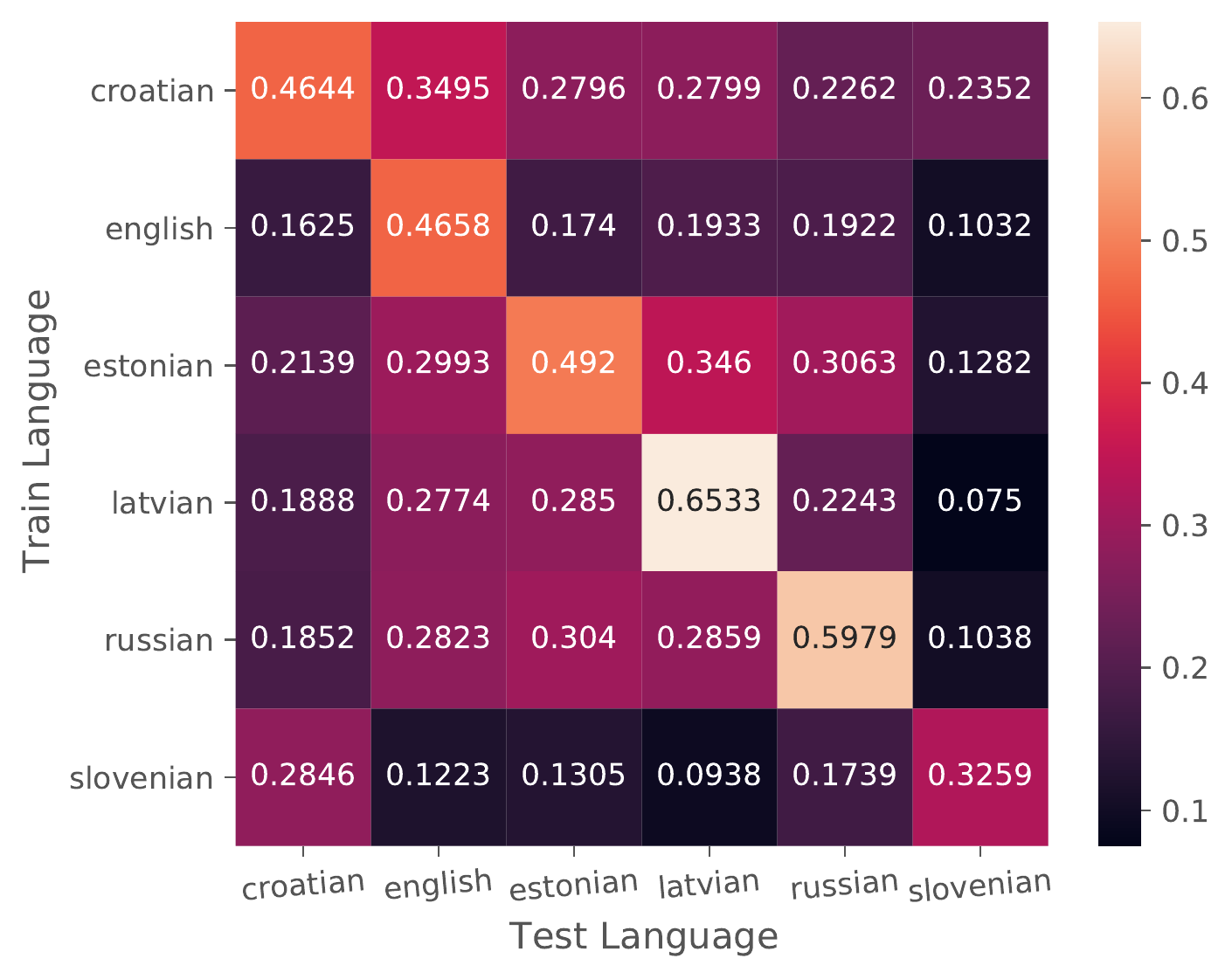}
    \caption{Evaluation of F1-score@10 retrieved keywords of the monolingual models in a setting of zero-shot cross-lingual learning. The rows represent the training language while the columns represent the testing languages. }
    
    \label{fig:heatmap}
\end{figure}

We explored how powerful are the monolingual (\textbf{MON}) models described in Section \ref{subsec:supervised} in a cross-lingual zero-shot keyword extraction setting. Each of six trained monolingual models was tested on six languages to obtain a heatmap presented in Figure \ref{fig:heatmap}. There was no single monolingual model that worked best for all of the remaining languages. For \textit{English}, the best-performing model was trained on \textit{Croatian}, most likely due to the fact that both datasets contain news from 2019, suggesting some topic intersection. The best performance on the \textit{Estonian} dataset was achieved by the model trained on the \textit{Latvian} dataset, most likely due to the fact that both of these datasets contain news from the same time period and were collected by the same media company, which covers news for both neighboring countries, Estonia and Latvia. Not surprisingly, the reverse correlation si also true: the best-performing model on the \textit{Latvian} dataset was trained on the \textit{Estonian} dataset. The best performance on the \textit{Russian} dataset was achieved by the \textit{Estonian} model due to both of the datasets coming from the same media-house stationed in Estonia, as reported by \newcite{koloski-etal-2021-extending}. Finally, the best performance on the \textit{Slovenian} data was achieved by the \textit{Croatian} model, most likely because both of these languages belong to the Southern-Slavic language group and since Slovenia and Croatia are neighbouring countries. 

\begin{figure}[h]
    \centering
    \includegraphics[width=\linewidth]{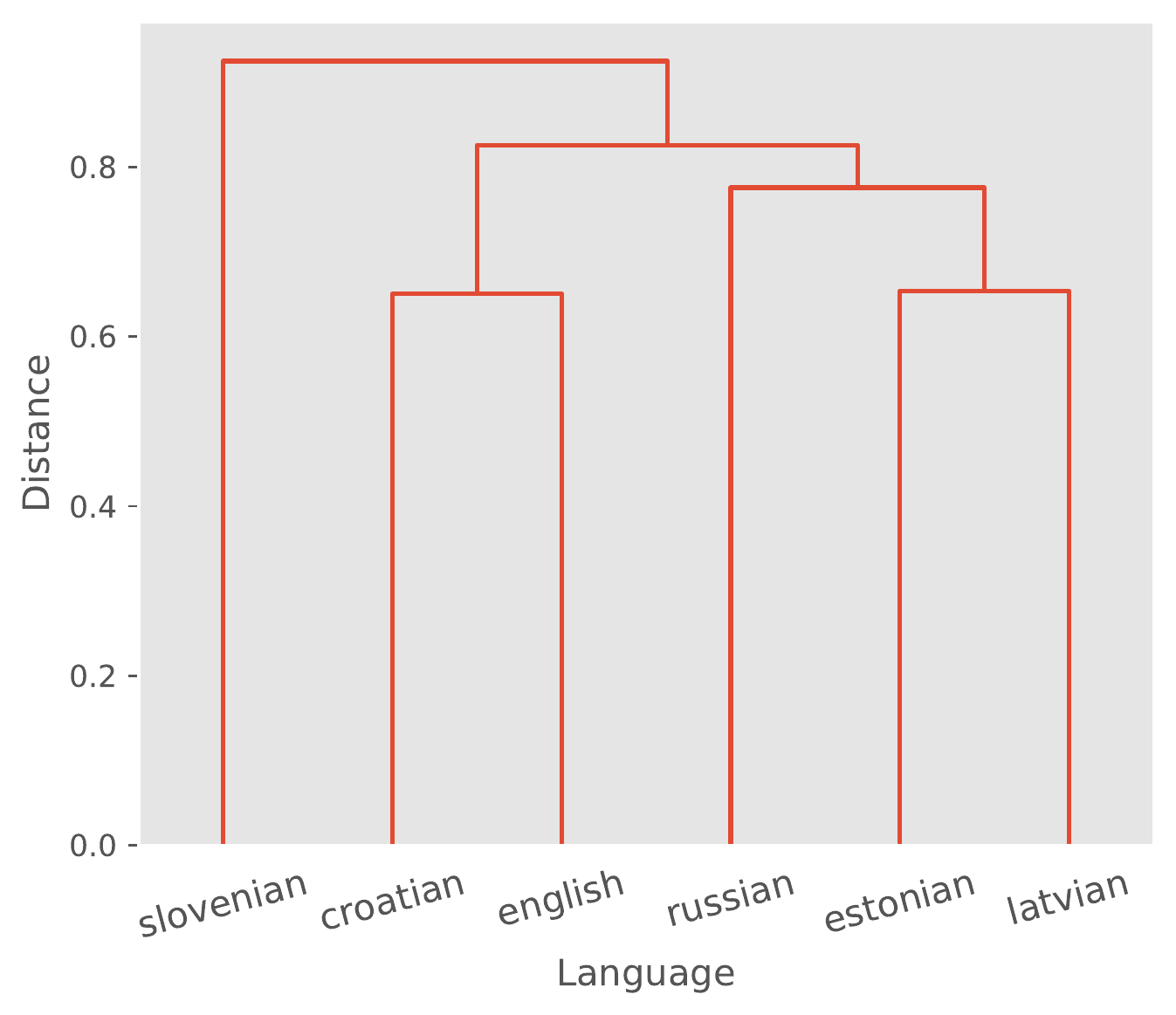}
    \caption{Dendrogram of the agglomerative clustering of the monolingual models applied in a cross-lingual setting.}

    \label{fig:hirearc}
\end{figure}

We also conducted hierarchical clustering, using the cross-lingual scores of the monolingual models as affinities. We present the resulting dendrogram in Figure \ref{fig:hirearc}. The results mostly confirm relations between languages, countries and sources of data, pointed out above. \textit{Estonian} and \textit{Latvian} datasets seem to be most similar. \textit{Russian} dataset is the natural addition to this cluster, most likely due to language and content similarity. Interestingly, \textit{Croatian} and \textit{English} form a separate cluster, most likely on the premises of both containing news from 2019, while the Slovenian dataset appears to be most dissimilar to other datasets. 

\section{Conclusions and Further Work}
\label{sec:conclusion}
In this work, we have presented a comprehensive comparison study covering multiple unsupervised, cross-lingual, multilingual and monolingual approaches for keyword extraction. While we did not manage to improve the performance of the supervised monolingual models by adding additional foreign language data to the training dataset, the results clearly indicate that cross-lingual models outperform unsupervised methods by a large margin. This suggests that if a labeled keyword train set from a specific domain is not available for a specific low-resource language, one opt to try to train a supervised model on a dataset covering the same domain in some other (preferably similar) language and employ that model in a zero-shot setting, before employing the unsupervised methods. 

While cross-lingual models tend to outperform unsupervised approaches by a large margin, the discrepancy in performance between the supervised cross-lingual setting and the supervised monolingual setting is nevertheless substantial and training the model on the target languages is still the preferred option in terms of performance. This is in line with further experiments conducted during the study, which suggest that the models perform really well for target languages similar to the languages on which the model was trained and when there is some intersection between the news content in the training and test datasets.

For further work we propose exploring few-shot shot scenarios, in which a small amount of target language data will be added to the multilingual train set. We plan to pinpoint the amount of needed target language data in order to bridge the gap in performance between the monolingual and cross-lingual models. Additionally, we propose ensambling multiple methods and explore how would that benefit the performance of the approach. 

\section{Acknowledgements}
This work was supported by the Slovenian Research Agency (ARRS) grants for the core programme Knowledge technologies (P2-0103), the project Computer-assisted multilingual news discourse analysis with contextual embeddings (CANDAS, J6-2581), as well as the European Union’s Horizon 2020 research and innovation programme under grant agreement No 825153, project EMBEDDIA (Cross-Lingual Embeddings for Less-Represented Languages in European News Media). The third author was financed via young research ARRS grant.
\section{Bibliographical References}\label{reference}

\bibliographystyle{lrec2022-bib}
\bibliography{lrec2022-example}

\label{lr:ref}
\bibliographystylelanguageresource{lrec2022-bib}

\end{document}